\documentclass[pdflatex,sn-mathphys,smallcondensed]{sn-jnl}

\jyear{2023}%
\usepackage{algpseudocode}

\usepackage{caption}
\usepackage{graphicx}
\usepackage{comment}
\usepackage{natbib}
\theoremstyle{thmstyleone}%

\usepackage{bm,graphicx}
\usepackage{amsmath}
\usepackage{fullwidth}
\usepackage{subcaption}
\usepackage{algorithmicx}
\usepackage{algorithm}

\usepackage{array}
\newcolumntype{P}[1]{>{\centering\arraybackslash}p{#1}}

\usepackage[inline, shortlabels]{enumitem}

\theoremstyle{thmstyletwo}%

\theoremstyle{thmstylethree}%

\begin{document}

\title{ARDDQN: Attention Recurrent Double Deep Q-Network for UAV Coverage Path Planning and Data Harvesting}


\author*[1]{Praveen Kumar}\email{praveen\_2221cs11@iitp.ac.in}
\author[2]{Priyadarshni}\email{priyadarshni\_2221cs16@iitp.ac.in}

\author[3]{Rajiv Misra}\email{rajivm@iitp.ac.in}
\affil[1]{\orgdiv{Department of Computer Science and Engineering}, \orgname{Indian Institute of Technology Patna}, \orgaddress{\state{Bihar}, \country{India}}}

\abstract{
Unmanned Aerial Vehicles (UAVs) have gained popularity in data harvesting (DH) and coverage path planning (CPP) to survey a given area efficiently and collect data from aerial perspectives, while data harvesting aims to gather information from various Internet of Things(IoT) sensor devices, coverage path planning guarantees that every location within the designated area is visited with minimal redundancy and maximum efficiency. We propose the ARDDQN (Attention-based Recurrent Double Deep Q Network), which integrates double deep Q-networks (DDQN) with recurrent neural networks (RNNs) and an attention mechanism to generate path coverage choices that maximize data collection from IoT devices and to learn a control scheme for the UAV that generalizes energy restrictions. We employ a structured environment map comprising a compressed global environment map and a local map showing the UAV agent’s locate efficiently scaling to large environments.

We have compared Long short-term memory (LSTM), Bi-directional long short-term memory (Bi-LSTM), Gated recurrent unit (GRU), and Bidirectional gated recurrent unit (Bi-GRU) as recurrent neural networks (RNN) to the result without RNN We propose integrating the LSTM with the Attention mechanism to the existing DDQN model, which works best on evolution parameters, i.e., data collection, landing, and coverage ratios for the CPP and data harvesting scenarios.
}


\maketitle

\section{Introduction}
\par With the rapid development of Unmanned Aerial Vehicles (UAVs) and next-generation wireless communication technologies, it is anticipated that the global market for UAV services will be worth USD 189.4 billion by 2030~\cite{1}. UAVs have become quite popular in various applications, including air-to-ground search, monitoring, mapping, search and rescue, precision agriculture, disaster management, aerial photography and videography, transportation and traffic management, and other aerial surveillance~\cite{2}. UAVs offer to liberate people from tedious, dangerous, and demanding task~\cite{3} while performing more efficiently than human vehicles. However, it has been challenging to cope with the demands of several real-world applications~\cite{4} because of the constrained IoT sensor range, speed calculations, and energy supply of a single UAV. This application needs an effective path-planning technique for efficient flying duration and obstacle avoidance management. One example of such an application is coverage path planning (CPP). Like conventional route planning, CPP aims to develop a path connecting the start and target locations when covering all places within the sphere of interest.

\par The main objectives of CPP are to locate NFZs, avoid barriers, and traverse as much of the target region as feasible while adhering to energy and path-length constraints. The number of sensing devices will expand 2.4 times, from $6.1\times 10^9 $ in 2018 to $14.7\times 10^{19}$ in 2023. By 2025, it is expected that the amount of sensed data will exceed 90 zettabytes due to the widespread deployment of these devices in several application scenarios~\cite{5}. This exponential spike shows the significance of vast sensing devices in collecting rich data. Data harvesting (DH), also known as data collection scenario, is the objective of the UAV to collect the data available at different IoT devices dispersed across an urban area. It is crucial to overcome difficult radio channel circumstances, such as alternating non-line-of-sight (NLoS) connections between the UAV and the IoT devices, owing to the barrier generated by buildings. Sensors are commonly used in scenarios including intelligent transportation, forest monitoring, and smart cities. UAVs with broad coverage and high mobility open up new possibilities for IoT data collection~\cite{6}.

\par The coverage ratio estimates the extent to which the UAV’s fly path covers a certain area. It can be defined as the ratio of the area covered by the UAV to the total area of interest. The landing ratio. The landing ratio measures the successful outcome of the UAV’s landings at particular locations or areas inside the covered area. It can be defined as the ratio of the number of successful landings to the total number of attempted landing points. The coverage ratio, landing ratio, and collection ratio are essential metrics used to evaluate the effectiveness of this approach in the context of CPP and data harvesting. 
Theile, Mirco, et al.~\cite{7} have examined coverage route planning and data harvesting problems with the help of a global map, local map concept, and the Double deep Q-network. Data collection from different sensors or IoT with the help of UAV based on the predefined area map has been considered by Bayerlein, Harald, et al.~\cite{8}, and has studied a similar concept of data harvesting~\cite{9} using multi-UAV. However, as far as I’m aware, no one has combined Deep Reinforcement Learning (DRL) and Recurrent Network (RNN) to improve the coverage, landing, and collection ratios for the CPP and data harvesting problem.

\par To improve the collection ratio, coverage ratio, and landing ratio for CPP and data harvesting problems using UAV, we have presented an Attention-based Recurrent Double Deep Q-Network (ARDDQN). The primary purpose of this work is to enhance DRL using RNN methods to generalize and extend UAV path planning issues and data gathering via CPP and data harvesting mechanisms. The network size, training parameters, and training duration will all rise if huge map sizes are used as direct inputs. The scalability difficulties of the traditional map-based input are resolved by combining RNNs with a global-local map approach. Path planning implies that nearby elements impact immediate actions like collision avoidance while distant factors influence overall path decisions. Consequently, the agent could learn more about close things than far-off ones. The global Map provides the agent with a broad description of all the Map’s features, and the agent is situated in the centre of a compacted version of the whole environment map. This representation yields a local map that contains comprehensive details about the agent’s immediate surroundings. The region pertinent to the UAV agent is then completely uncompressed and clipped from this local Map.

\par The overall structure of this work is described as follows: The related work is described in Section II. The system model is explained in Section III. Section IV shows the experimental setup, followed by a discussion of the findings—Section V is the conclusion of the work.

\section{Related Work}
\par Although several path-planning approaches for CPP difficulties have been put forward, employing RNN in conjunction with DRL gives a chance to approach the issue more effectively. DRL agents can learn control techniques that generalize over various scenario parameters when the scenario changes, eliminating the need for costly re-training or re-computation. The productivity of DRL inference in terms of computing, the complexity of problems involving autonomous UAV control, the prevalence of frequently non-convex optimization problems that have been demonstrated to be NPhard in many cases, and the DRL paradigm’s adaptability to prior knowledge and environmental assumptions are all factors that contribute to its popularity in this context. Zeng et al.~\cite{10} provide an overview of the problems associated with UAV use in communication networks, particularly IoT data collection. In prior UAV route planning investigations, convolutional map processing for DRL agents has already been applied in~\cite{11}. A double deep Q-network agent provides the local significance of the monitoring map region, depicting the agent’s neighborhood cropped to a specific size in the drone patrolling challenge outlined in~\cite{12}. There is no consideration for physical environment information or navigational constraints like the total time of flight collision avoidance.

\par Fixed-wing UAVs are untrusted with keeping an eye on a fire that is spreading in a stochastic manner over time. Assessments or belief maps given to the Deep reinforcement learning agents inform control decisions. Here, the emphasis is on the problem’s high instability rather than juggling navigational limitations and mission objectives in vast, complicated landscapes. The quadcopter UAVs~\cite{13}, which are set in a comparable environment without navigational restrictions and use uncertainty maps to help path planning, likewise have the task of monitoring wildfires as their primary mission. The extended Kalman filter is the foundation of their strategy rather than the RL paradigm. 

\par Baldazo et al.~\cite{14} propose a multi-agent Deep reinforcement learning technique for flood monitoring that uses local observation of the inundation map from the UAVs to make decisions in real-time while monitoring another natural catastrophe scenario. All the strategies that have been discussed focus on dealing with a particular type of UAV flight in uncomplicated physical settings, but none of them consider combining regional and global map data. The study~\cite{15} provides a vital instance of using UAVs to deliver communication services to land users or equipment. The research utilizes a path-planning technique for UAVs that uses map data and focuses on a difficult urban setting. This method blends sequential and dynamic convex programming approaches to optimize the UAV’s trajectory in the current situation. A compact map with obstructions is used in~\cite{16} for concurrent data collection by ground and airborne vehicles. The entire global map data may be given to the DRL agents because of the tiny size of the Map. Zhang et al.~\cite{17} investigate a scenario where a UAV linked to a cellular network must use a radio map to move from one location to another while maintaining contact with a terrestrial network. The approach is not RL-based and does not use more precise local maps or strict navigational restrictions.

\par The following are this paper’s key contributions:
\begin{itemize}
    \item Recurrent neural networks are added to the existing DDQN-based model and global-local map data, introducing a unique method that enables DRL to achieve better CPP and DH.
    \item Comparison of performance of different RNN architectures, namely long short-term memory (LSTM), Bidirectional long short-term memory (Bi-LSTM), Gated recurrent unit (GRU), and Bidirectional gated recurrent unit (Bi-GRU), with a base model (DDQN Network). 
    \item Primarily, we propose an Attention-based Recurrent Double Deep Q-Network (ARDDQN), where we focus on using LSTM as a recurrent Network to get the best model performance.
    \item Improved data collection in DH scenarios and significantly improved landing and coverage rates in CPP scenarios.
\end{itemize}

\section{System model}

\begin{figure*}[ht!]
  \includegraphics[width=\textwidth]{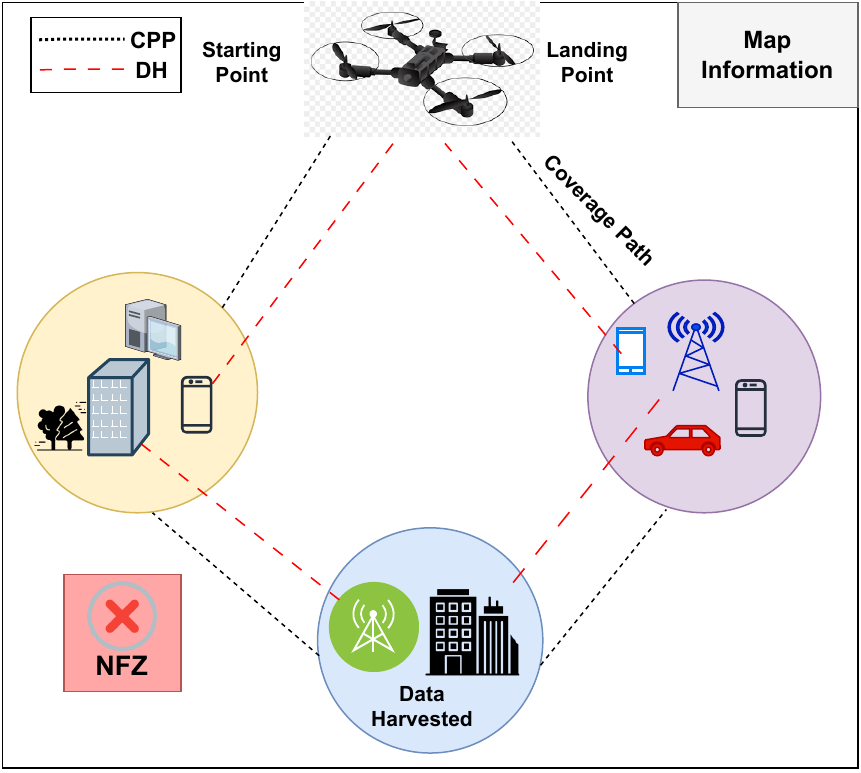}
  \caption{Proposed scenario for UAV-assisted coverage path planning and data harvesting}
  \label{fig:systemmodel}
\end{figure*}

Figure~\ref{fig:systemmodel} represents the  UAV-assisted coverage path planning and data harvesting. In this diagram, a dynamic and complex scene is shown through the map information as a drone takes off from its predetermined starting position called starting point. The concept of ``Path Coverage" is illustrated by the thorough plotting of its course over the map. The drone communicates with a variety of Internet of Things (IoT) devices that are positioned strategically along its path in order to obtain information. Through these contacts, the drone is able to harvest useful data. The map highlights certain areas as ``Non-Flying Zones" indicated in red, to guarantee a seamless and obstacle-free flight. These areas indicate restricted zones or probable obstacles that the drone must avoid while flying. The map includes a ``Landing Point". The drone has been programmed to arrive at this location when its mission is over or its battery is about to drain out.

\subsection{Environment Model }
\par We take the environment as the grid size in square shape $ G\times G \in \mathbb{N}^2$ with a cell with size c, the set of available potential positions $\mathcal{G}$ and $ \mathbb{N} $ representing the set of natural numbers. 
Our proposed model applies to any environment with a square-shaped grid representation of any real-world area. The environment must be discretized before using our map-processing approach. Regulation No-Fly Zones (NFZs), defined start/landing spots, and impediments may all be found in the environment. 
$ {\mathbf{L}} $ number of designated starting /landing position for the environment is given as: 


\begin{equation}
\widetilde{\mathbf{L}}=\left \{{ \left [{x^{l}_{i},y^{l}_{i}}\right ]^{\mathrm {T}}, i = 1,2,...,{\mathbf{L}}, \left [{x^{l}_{i},y^{l}_{i}}\right ]^{\mathrm {T}} \in \mathcal {G}}\right \}
\label{e:eq1}
\end{equation} 
\par A set that provides a combination of ${\mathbf{Z}}$ number of places that the UAV cannot occupy is indicated as:

\begin{equation}
\widetilde{\mathbf{Z}}=\left \{{ \left [{x^{z}_{i},y^{z}_{i}}\right ]^{\mathrm {T}}, i = 1,2,...,{\mathbf{Z}}, \left [{x^{z}_{i},y^{z}_{i}}\right ]^{\mathrm {T}} \in \mathcal {G}}\right \}
\label{e:eq2}
\end{equation} 

\par Equation~\ref{e:eq2} includes tall buildings that UAVs cannot fly over and authorized no-fly Zones (NFZ). The set provides the number of ${\mathbf{B}}$ obstructions blocking wireless networks is referred to as:

\begin{equation}
\widetilde{\mathbf{B}}=\left \{{ \left [{x^{b}_{i},y^{b}_{i}}\right ]^{\mathrm {T}}, i = 1,2,...,{\mathbf{B}}, \left [{x^{b}_{i},y^{b}_{i}}\right ]^{\mathrm {T}} \in \mathcal {G}}\right \}
\label{e:eq3}
\end{equation} 

\par All buildings, including the tiniest ones that UAVs may fly above, are included in the collection. Lowercase letters l, z, and b may be used to indicate the coordinates of elements of the environment using $\widetilde{\mathbf{L}}$,$\widetilde{\mathbf{Z}}$ and $\widetilde {\mathbf{B}}$.

\subsection{UAV Model }
\par The UAV occupies a single cell while moving across the environment at a constant height h. The following attributes define the UAV’s state.

\begin{itemize}
    \item UAV Position: $\mathbf{P}(t)=\left[x(t), y(t), z(t)\right]^T \in \mathbb{R}^3$ with height $z(t) \in\{0, h\}$, it can be at ground position or at constant height $h$;
    \item Status of operation: $\theta(t) \in [0,1] $ it can be active (1) or inactive (0) mode;
    \item  Energy level of battery: $E_b (t)\in \mathbb{N}$, after performing one action step, it will be decremented by 1.
\end{itemize}

\par Combining the UAV agent’s flight time and height will simplify incorporating the altitude in the observation space. Due to the environment’s high-rise building density, however, this approach only considers 2D trajectory optimization. The analysis does not consider that flying over these buildings would require long ascending periods. The UAV’s onboard battery capacity limits the mission duration, which lowers the effectiveness of 3D control for collecting data. At the end of every operation, UAVs must be successfully landed at their ground position because upward flight consumes more energy~\cite{18}. These limitations limit the effectiveness of data collecting as a whole. A UAV’s data-collection task concludes after $T\in N$ task time steps. The time horizon is separated into equal mission time slots, indicated by the notation $t\in [0, T]$, with each mission time slot lasting for $\nu_t$ seconds. The UAV’s action space with cell size c and height h can be defined as follows~\cite{8}:

\begin{equation} \mathcal {A_s}=\left \{{ \underbrace {\begin{bmatrix}0\\ 0\\ 0\end{bmatrix}}_{\text {Hover}}, \underbrace {\begin{bmatrix}c\\ 0\\ 0\end{bmatrix}}_{\text {East}},\underbrace {\begin{bmatrix}-c\\ 0\\ 0\end{bmatrix}}_{\text {West}}, 
\underbrace {\begin{bmatrix}0\\ c \\ 0\end{bmatrix}}_{\text {North}},\underbrace {\begin{bmatrix}0\\ -c\\ 0\end{bmatrix}}_{\text {South}}, \underbrace {\begin{bmatrix}0\\ 0\\ -h\end{bmatrix}}_{\text {Land}} }\right \}
\label{e:eq4}
\end{equation}

\par The movement actions of the UAV $a(t) \in  {\mathcal {B}}\left ({\mathbf {P}(t)}\right ) $, which is describe as follows: 

\begin{equation} {\mathcal {B}\left ({\mathbf{P}(t)}\right )=\begin{cases} \mathcal {A_{s}}, & \mathbf {P}(t) \in \widetilde{\mathbf{L}}\\ \mathcal {A_{s}} \setminus [0,0,-h]^{T}, &\text {otherwise},\end{cases}}
\label{e:eq5}
\end{equation}

\par All feasible actions are $\mathcal{A_{s}}$. In particular, landing operations are only permitted when the UAV is in the landing zone. The UAV covers a distance equal to the cell size c in each time slot. Since the mission time slots are selected to be suitably brief, we may assume that the UAV’s velocities $v(t)$ remain constant throughout a single time slot. The UAV must move vertically at a speed of $V=c/\nu_t$ or cease to move horizontally, i.e., $v(t)\in [0, V] $ for every $t\in [0, T]$. According to the motion model offered by the UAV, the location of the UAV varies as:

\begin{equation}
     \mathbf {P}(t+1)=\begin{cases} \mathbf {p}(t) + \mathbf {a}(t), & \theta(t)=1\\ \mathbf {p}(t), & \text {otherwise}, \end{cases}
     \label{e:eq6}
\end{equation}

\par The UAV will be stationary when it is not in use. The UAV operational state changes according to the following expression:

\begin{equation}
\theta(t+1)=\begin{cases} 0, & \phantom {\lor }\mathbf {a}(t) = [0,0,-h]^{T} \\ & \lor \theta(t) = 0\\ 1, & \text {otherwise},\\ \end{cases}
\label{e:eq7}
\end{equation}

\par Finally, the operational status will turn inactive after the UAV landed successfully. The moment when the UAV achieves its terminal state and stops actively working, or when the operational state of all UAVs is $\theta(t)=0$, is known as the end of the data harvesting and coverage path planning mission. The UAV battery content changes according to the following:

\begin{equation} 
E_b\left ({t+1}\right )=\begin{cases} E_b(t) - 1, & \theta(t) = 1\\ E_b(t), & \text {otherwise},\\ 
\end{cases}
\label{e:eq8}
\end{equation}

\par Based on the assumption that the UAV would consume energy continuously while it is in use and won’t use any once it has stopped. This simplification is acceptable because the hovering component consumes most of the power for UAVs.

\subsection{Communication Channel Model}
\par The conception of a communication time slot, defined as $n \in [0, N]$, where $N = \alpha T$, is introduced to describe the communication between UAV and the IoT sensor nodes available in the target area. This is not the same as the task time slot. Each mission time slot has enough communication time slots allotted to it such that the position and channel gain of the UAV may be considered constant throughout a single communication time slot. According to~\cite{19}, the length of a communication time slot, n, is $\nu_{n}=\nu_t/\alpha$ second. The location of the kth sensor node is $\mathbf{U}_k=\left[x_k, y_k, 0\right]^{\mathbf{T}} \in \mathbb{R}^3$, where $k\in K$, the total number of sensor nodes. These sensor nodes are equidistant from one another and have fixed coordinates for the whole mission time slot T. The kth sensor node is given a certain range of random data ${\mathcal{I}_k(t=0)}={\mathcal{I}_{k, init}}$ at the start of the mission. We define that the kth sensor node creates new data $\mathcal{I}_{k, n e w} \sim$ Poisson $\left(\lambda_p\right)$ according to the standard Poisson distribution expression.

\par A UAV-to-ground channel model is used to analyze the communications between the UAVs and the sensor nodes~\cite{8}. In this approach, the kth sensor node’s maximum attainable data rate during communication time slot $n$ is specified as:

\begin{equation}
{\mathcal{R}}_{k}(n)=\log _2\left(1+\mathrm{SNR}_{k}(n)\right)
\label{e:eq9}
\end{equation}

\par It is important to notice that the length of a communication time slot is adjusted to ensure that the maximum attainable data rate stays constant regardless of the communications volume within a mission time slot. It is possible to express the information’s effectiveness rate as the effective information rate is expressed as $\mathcal{I}_k(n)$ given the amount of data present at the $kth$ device.

\begin{equation} 
\mathcal {R}_{k}(n)=\begin{cases} \mathcal {R}_{k}(n), & \mathcal{I}_{k}(n) \geq \nu _{n}\mathcal {R}_{k}(n) \\ \mathcal{I}_{k}(n)/ \nu _{n}, & \text {otherwise}. \end{cases}
\label{e:eq10}
\end{equation}

\par The SNR is defined as the ratio of the transmit power ${\mathcal{P}_{k}}$ to the white Gaussian noise power at the receiver ${\sigma ^{2}}$, the distance between the UAV and the device ${d}_{k}$, the path loss exponent $\alpha _{e}$, and the Gaussian random variable $\eta_{e}\sim {\mathcal{N}}(0,{\sigma ^{2}})$: 

\begin{equation}
{SNR}_{k}(n) = \frac {\mathcal{P}_{k}}{\sigma ^{2}} \cdot {d}_{k}(n)^{-\alpha _{e}} \cdot 10^{\eta _{e}/10} 
\label{e:eq11}
\end{equation}

\par where $e \in {[LoS, NLoS]}$. Since Equation~\ref{e:eq11}  represents the signal-to-noise ratio average over low fading, B’s collection of obstacles in the urban environment strongly impacts the propagation factors on the line of sight and non-line of sight. UAVs serve sensor nodes using simple time division multiplexing (TDMA). Here, for each communication time slot $n \in [0, N ]$,  sensor node $k\in [1, K]$ has the highest SNR$_K$(n). A planning algorithm selects the files to be uploaded. The TDMA constraint for scheduling variables $q_{k}(n)\in[0, 1]$ is given as:

\begin{equation}
\sum_{K=1}^{K}q_{k}(n)\le1, n\in [0,N]
\label{e:eq12}
\end{equation}

\par The achievable throughput of a mission time slot t is the sum of the achieved rates of the corresponding communication time slots $n \in [\delta t, \delta(t + 1) $-$ 1]$ on K sensor nodes and is given as:

\begin{equation}
C(t)=\sum_{n=\delta t}^{\delta(t+1)-1}\sum_{k=1}^{K}q_{k}(n)R_{k}(n)
\label{e:eq13}
\end{equation}

\subsection{Definitions of the Mission and Targets}
\subsubsection{Coverage Path Planning}
\par The problem with coverage that has to be rectified may be shown with a grid map consisting of two dimensions and three channels. Each cell in the grid represents a square region of the coverage zone. The three channels provide descriptions of the starting and landing zones (blue), target zones (green), and no-fly zones (red). After completing a coverage path, the agent can begin and land in the start and landing zones. The UAV camera’s field of view (FoV) must pass over the target zones at least once. No-fly zones (red) are places where drones cannot fly. The critical point is that the takeoff and landing zone can not be part of a no-fly zone. Coverage routing aims to pass over or relatively near a chosen target area to be visible to a camera sensor connected to the underside of the UAV. The target area may be defined by $T_{trg}(t) \in B^{G \times G} $, where each element signals whether or not a cell has visited. It is possible to determine the camera’s field of vision at present using $V_{ew}(t)\in B^{G \times G}$ signaling for each cell whether or not it is covered in the present field of vision. In this work, we assume the current UAV location is surrounded by a $5\times 5$ field of vision. Buildings may also obscure a person’s line of sight, which is considered when determining $V_{ew} (t)$. As a consequence, the UAV is unable to view around corners.

\begin{equation}
T_{rg}(t+1)=T_{rg}(t) \wedge \neg V_{ew}(t)
\label{e:eq14}
\end{equation}

\par Takeoff, landing zones, and no-fly zones are allowed to be coverage targets under our mission specification. However, obstacle cells in the environment are not. It would be best to fly over as much of the target region as possible in the allowed time.

\subsubsection{Data Harvesting}
\par While each IoT device’s location is represented by $u_k\in \mathbb{N}^2 $, the goal of planning a path for wireless data collection is to gather data from $K\in \mathbb{N}$ stationary IoT devices dispersed across the environment at ground level. The quantity of information that must be collected from each device by the UAV is $\mathcal{I}_k(t) \in R$. A typical extended distances path loss model with Gaussian shadow fading is used to calculate the data throughput $Th_k(t)$ between the chosen device k and the UAV, depending on whether a line-of-sight connection can be made or obstructions are present. When communicating, the UAV identifies the device with the most data and the quickest data rate. \cite{13} thoroughly explains the connection performance and multiple access mechanisms. Every device’s data changes as:

\begin{equation}
I_k(t+1)=I_k(t)-Th_k(t)
\label{e:eq15}
\end{equation}

\subsubsection{Unifying Map-Layer Description}
\par Every cell contains a device, with the exception of the landing/ starting zones and obstructions. The aim of the data harvesting challenge is to extract the maximum amount of data from the devices within the designated flight duration.One target layer on the map $\mathcal{D}(t)\in R^{G\times G}$ may be used to explain both issues. The target map layer in CPP is provided by $T_{rg}(t)$, developing per Equation~\ref{e:eq14}. The cell at location $I_k$ has value $\mathcal{D}(t)$ and is developing under Equation~\ref{e:eq15} in DH, where the target map layer shows how much data is accessible in each cell that one of the gadgets is occupying. A cell’s value is zero if it doesn’t have any devices or if all of the device data has been gathered. Both issues can be addressed using deep reinforcement learning and a neural network with the same topology since the state representations used to describe the two problems are identical.

\subsection{Partially Observable Markov Decision Process}
\par Using a partly observable Markov decision process (POMDP), the problems mentioned above are solved. A tuple (${S_p}, \mathcal{A}, Z, \mathcal{R}, T, O, \gamma)$ is used to describe the POMDP, where $ {S_p}$ indicate the state space, $\mathcal{A}$ describe the action space, and $Z: {S_p} \times  \mathcal{A} \times{S_p} \rightarrow \mathcal{R} $ mapped the probability function for the transition  The reward function is shown as,$\mathcal{R}:{S_p}\times \mathcal{A} \times {S_p}\rightarrow \mathcal{R} $, also provides a real-valued reward for each state, action, and subsequent state triplet combination. $ O:{S_p}\rightarrow T $  denotes the observation function, and $T$ defines the observation space. The discount factor($\gamma$) determines the significance of long-term and short-term benefits and $\gamma\in [0,1]$ values. The state of the UAV coverage path planning can be described as follows:

\begin{multline}
{S_p}=\underbrace{\mathbb{B}^{G\times G \times 3}}_{\begin{array}{c}
\text {Given Environment} \\
\text {Map }
\end{array}} \times \underbrace{\mathbb{R}^{G \times G}}_{\begin{array}{c}
\text {Target area} \\
\text { Map }
\end{array}} \\
\times \underbrace{\mathbb{N}^2}_
{\text {Position of the UAV}} \times 
\underbrace{\mathbb{N}}_{\substack{\text { UAV Flying } \\
\text {Time }}}
\label{e:eq16}
\end{multline}

\par The element of state space $s(t)\in {S_p}$ given as: 

\begin{equation}
s(t)=({G}, \mathcal{D}(t), \mathbf{P}(t), E_b(t))
\label{e:eq17}
\end{equation}

\par The four tuples of state space element are as follows:
    \begin{itemize}
        \item  G as environment map contains all information like takeoff, landing, obstacles, and no-fly zone.
        \item $\mathcal{D}(t)$ as the target map showing remaining information at device locations or cells that will be exposed at time t
        \item P(t) shows the UAV location at a particular time t
        \item $E_{b}(t)$ is the battery’s energy level for the UAV at time duration t.
    \end{itemize}

\par The UAV’s action space, $a(t)\in \mathcal{A_s}$, is specified as $\mathcal {A_s} = [N, E, S, W, H, L]$. Here, the four directions are N, E, S, and W. H and L are the UAV’s hovering and landing actions. The following components make up the generic reward function $\mathcal{R}(s(t), a(t), s(t + 1))$ as:

\begin{itemize}
\item $r_c$ (+ve): The difference between s(t + 1) and s(t) in the quantity of newly covered target cells or the cell coverage or data collecting reward made possible by the acquired data.
\item $r_sc$(-ve): penalty for the safety controller (SC) if the drone must be stopped from hitting an obstacle or entering an NFZ.
\item $r_mov$ (-ve): Constant mobility penalty incurred for each action the UAV does that does not result in the mission’s completion
\item $r_{crash}$ (-ve): Penalty if the drone does not safely land in a landing zone before its remaining flight time expires
\end{itemize}

\subsubsection{Map Processing}
\begin{figure*}[ht!]
  \includegraphics[width=\textwidth]{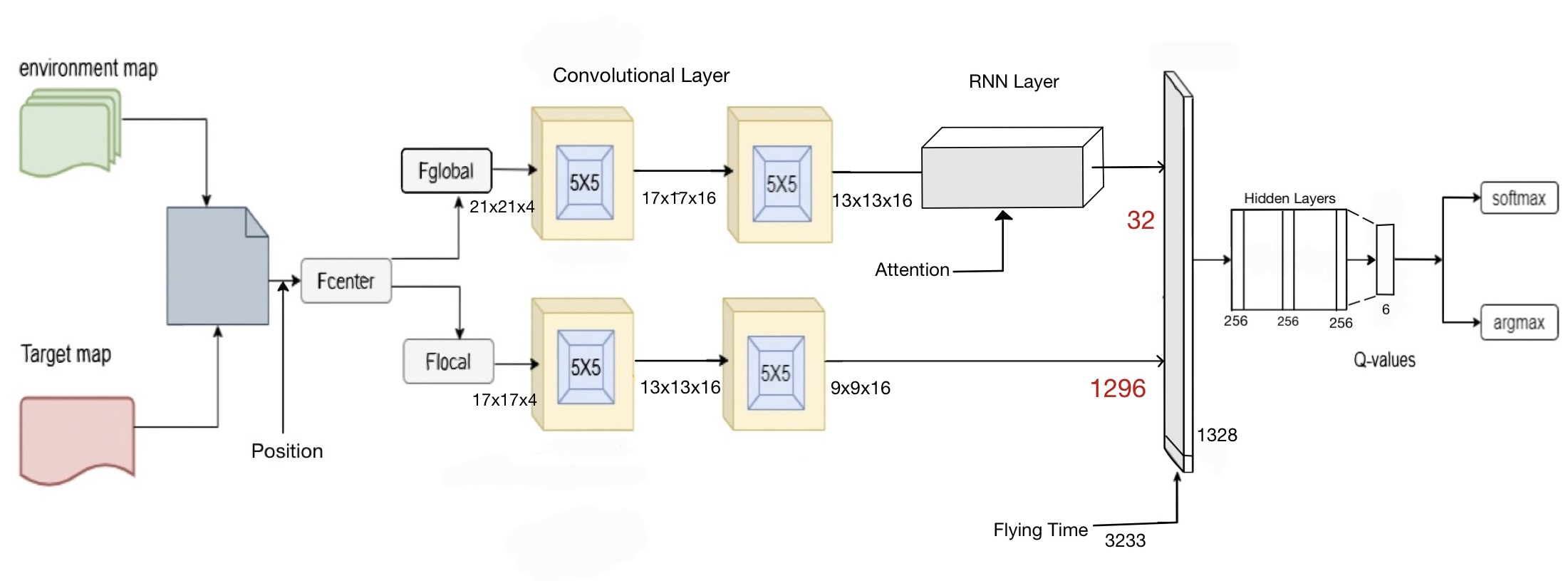}
  \caption{Figure shows the proposed model for map processing.}
  \label{fig:figure1}
\end{figure*}

Two map processing techniques help the agent comprehend the vast state space. The first approach centers the Map around the agent, considerably boosting the agent’s efficiency. The disadvantage of this strategy is that it enlarges the state space representation. The centered Map is used as two inputs in the second map processing stage: a full-depth local map concentrating on the agent’s local area and a compressed global map offering a more comprehensive but less detailed perspective. Figure~\ref{fig:figure1} shows the data flow demonstrating the use of these features.

\subsubsection{Double deep Q-Network}
\par We employ one model-free and off-policy-based reinforcement learning mechanism, double deep Q-networks (DDQNs), to resolve the above formulated POMDP suggested by~\cite{20}. The DDQNs generally represent the Q-value of each state-action combination as:

\begin{equation}
\mathcal{Q}^\pi(s(t), a(t))=\mathcal{E}_\pi\left[\sum_{k=t}^T \gamma^{k-t} \mathcal{R}(s(k), a(k), s(k+1))\right]
\label{e:eq:18}
\end{equation} 

\par This procedure reaches the best Q-value. The first Q-network is updated by reducing the loss caused by experience replay memory in two Q-networks parameterized by $\phi$ and $\bar \phi$.

\begin{equation}
Loss (\phi)=\mathcal{E}_{s, a, s^{\prime} \sim \mathcal{D}}\left[\left(\mathcal{Q}_\phi(s, a)-\mathcal{Y}\left(s, a, s^{\prime}\right)\right)^2\right]
\label{e:eq:19}
\end{equation}

\par The target values can be represented as:

\begin{equation}
\mathcal{Y}\left(s, a, s^{\prime}\right)=r(s, a)+\gamma \mathcal{Q}_{\bar{\phi}}\left(s^{\prime}, \operatorname{argmax} \mathcal{Q}_\phi\left(s^{\prime}, a^{\prime}\right)\right)
\label{e:eq:20}
\end{equation}

\par The second network parameter can be updated through the following equation:

\begin{equation}
\bar{\phi} \leftarrow (1-\eta) \bar{\phi}+\eta \phi
\label{e:eq:21}
\end{equation}

Figure~\ref{fig:figure1} illustrates the two Q-networks' neural network architecture. Upon layering and centering around the UAV's position, the environment map and target map are converted into components for both local and global observation, respectively. Using ReLU activation functions, the produced tensors are then flattened, concatenated, fed through two convolutional layers of each, fed through an RNN, and ultimately fed through three hidden layers. RNNs are the most effective at extracting temporal information. The main objective is to optimize CPP and DH tasks using the temporal and spatial data from the UAV. We focused on careful attention to the tests of LSTM, Bi-LSTM, GRU, and Bi-GRU. Whether the Q-values are transmitted via an argmax function to form an exploitation distribution or a softmax function to produce an active distribution for exploration, the output layer without an activation function displays the Q-values directly.

\subsection{Recurrent Neural Network}
\par RNN's recurrence implies that it does the same task for each sequence by being dependent on the preceding step~\cite{21}. It has two fundamental features: 

    \begin{itemize}
        \item a hidden state that is distributed in nature and allows for storing past information and
        \item the ability for hidden states to update themselves in complex and nonlinear dynamics.
    \end{itemize}

\par To solve the problems that occur during the training of vanilla RNNs, the following modifications were introduced:

\begin{itemize}
    \item The LSTM network proposed by Hochreiter et al.~\cite{22} is an extension of RNNs that has been redesigned to address RNN vanishing and exploding issues.
    \item GRU is an LSTM branch introduced by Cho et al.~\cite{23}. It is a replica but excludes an output gate that lets content flow from the memory cell to the large net at each time step. It is said to be fast for training purposes, and its internal makeup is not complex but requires a few computations to update the hidden layer.
    \item The Bi-LSTM network, based on the LSTM, combines input sequence information in both the forward and backward directions to extract context information better.
    \item The Bi-GRU network is similar to that of Bi-LSTM, except for the cyclic unit. These two bidirectional networks can employ forward and backward information at the same time~\cite{24}. 
\end{itemize}

\subsection{ARDDQN Architecture}
\par The structure of this work mainly focuses on the Double Deep Q-Network (DDQN) approach. Additionally, it introduces an attention-based Long Short-Term Memory (LSTM) network.

\subsubsection{Long Short Term Memory}
\par LSTM is suitable for analyzing and predicting critical events with very long intervals in time series because of its unique architectural structure~\cite{25}. Each LSTM block also includes a memory cell and three gates that regulate the flow of information to its cell state $c_t$ with an input gate $i_t$, a forget gate $f_t$ and an output gate $o_t$~\cite{26}. Each of the three gates also performs a distinct function. The forget gate determines which information is discarded; the input gate controls which information is fed into the cell state, and the output gate controls the LSTM cell’s outgoing information~\cite{27}. The values for the gates, candidate cell state $\widetilde{c_{t}}$ and actual cell state $c_t$ at time t are computed as follows based on the input x$_t$ and the previous output h$_{t-1}$:

\begin{equation}
f_t = \sigma(W_f x_t + U_f h_{t-1} + b_f)
\end{equation}
\begin{equation}
i_t = \sigma(W_i x_t + U_i h_{t-1} + b_i)
\end{equation}
\begin{equation}
o_t = \sigma(W_o x_t + U_o h_{t-1} + b_o)
\end{equation}
\begin{equation}
\widetilde{c_{t}} = \tanh(W_{\widetilde{c}} x_t + U_{\widetilde{c}} h_{t-1} + b_{\widetilde{c}})
\end{equation}
\begin{equation}
c_t = f_t c_{t-1} + i_t \widetilde{c_{t}}
\end{equation}

\par where $x = (x_1, . . . , x_n)$ is the input vector where $x_{t},$ (t = 1,..., n) is the data point at time t in a sequence of length n. $c_t$ is the cell state, i.e., the memory of the cell at time t. $\widetilde{c_{t}}$ is the candidate cell state. $h_{t}$ is the output of the cell, also called the hidden state. $f_{t}, i_{t},$ and $o_{t}$ are the values for the forget, input, and output gates, respectively. $W_{f}, W_{i}, W_{o},$ and $ W_{\widetilde{c}}$ are the weight matrices associated with the input x. $U_{f}, U_{i}, U_{o}, $ and $U_{\widetilde{c}}$ are the weight matrices associated with the output $h_{t}$. $b_{f}, b_{i}, b_{o}, $ and $ b_{\widetilde{c}}$ are the bias vectors. Finally, the output h$_t$ is computed as:

\begin{equation}
h_t = o_t \tanh(c_t)
\end{equation}

\subsubsection{Attention based LSTM}
\par The majority of attention models use the Encoder-Decoder framework. In the Seq2Seq model, the encoding process of the original encoder-decoder model generates an intermediate vector, which is utilized to hold the information of the original sequence~\cite{28}. The length of this vector, however, is fixed. When the input original sequence is long, this vector can only hold a limited amount of information, limiting the model’s comprehension capabilities. Using the attention method to overcome the original codec model’s limits on fixed vectors~\cite{29}. The following steps primarily implement the attention mechanism. The LSTM output $[h_1, h_2, h_3,...., h_n]$ is non-linearly converted to obtain $[u_1, u_2, u_3,...., u_n]$. Some operating parameters greatly influence the attitude and location of shield tunneling during the process. Thus, these parameters should be given more weight. The attention mechanism generates the attention weight matrix [1, 2, 3,...., n], representing each intermediate condition’s significance. Finally, the encoding vector V is obtained by performing a weighted sum of the input parameter and weight~\cite{10}. Decoding according to the encoding vector V yields the output y, which is as follows:

\begin{equation}
u_k = \tanh(W_kh_k+b_k)
\end{equation}
\begin{equation}
a_k = \frac{\exp(u^T_ku_s)}{\sum_{i=1}^n \exp(u_i)} 
\end{equation}
\begin{equation}
V = \sum_{k=1}^n a_k h_k
\end{equation}

\par where $u_k$ is the unnormalized attention score for the $k$th head. $W_kh_k$ is the weighted key-query product for the $k$th head. $b_k$ is the bias vector for the $k$th head. $a_k$ is the attention weight for the $k$th head. $u^T_ku_s$ is the unnormalized self-attention score for the $k$th head. $u_i$ is the unnormalized self-attention score for the $i$th head. $K$ is the number of heads. $h_k$ is the hidden state for the $k$th head, i.e, output of LSTM. $V$ is the attention output.

\begin{figure*}[ht!]
  \includegraphics[width=\textwidth]{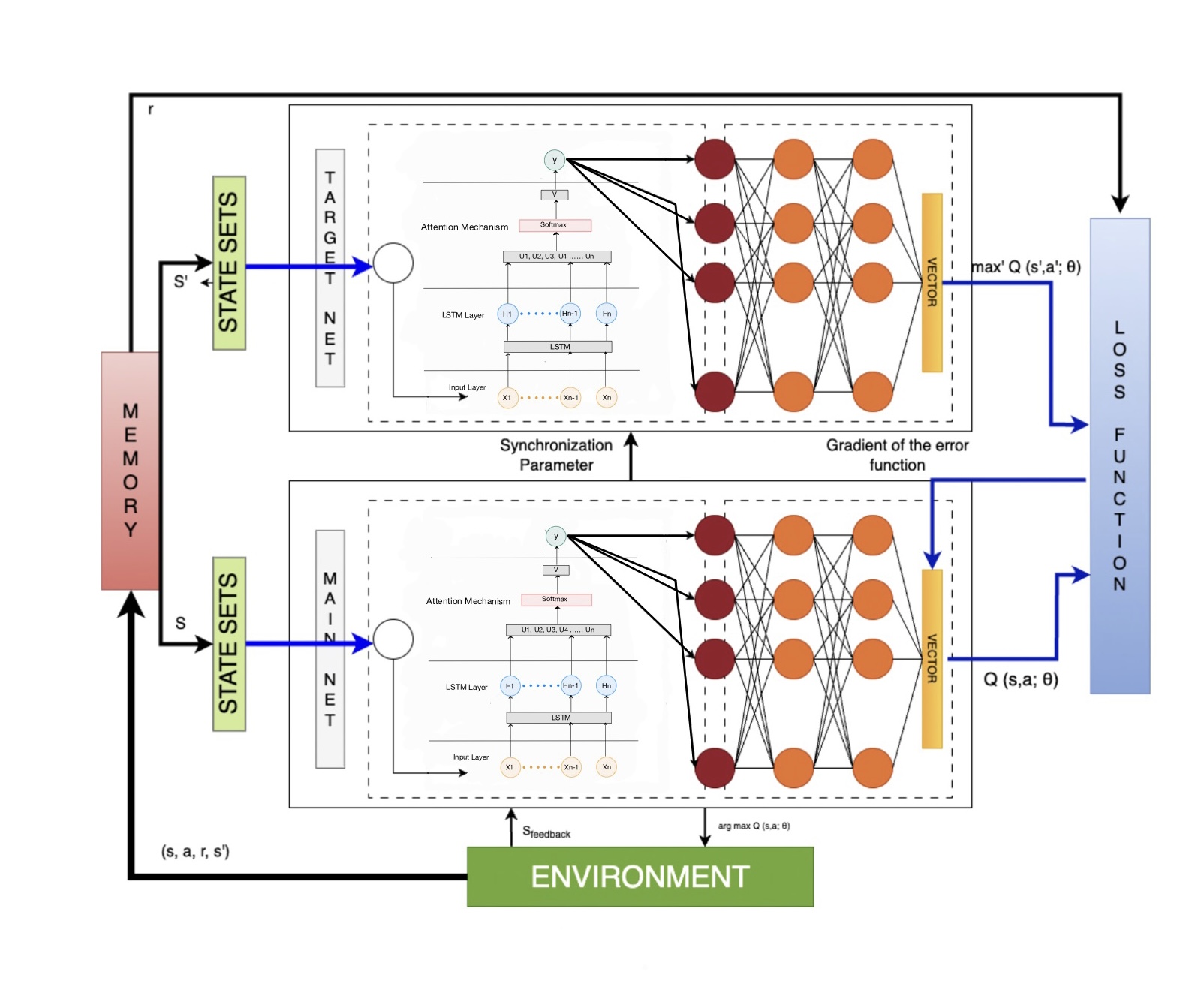}
  \caption{Figure shows the proposed integrated architecture for attention-based recurrent networks with the DDQN Model.}
  \label{fig:figure2}
\end{figure*}

\subsubsection{Architecture summary}
\par The LSTM network captures and stores pertinent information from the input states within its cell units. Illustrated in Figure~\ref{fig:figure2}, the network comprises two interconnected components: 

\begin{enumerate}
\item Primary(main) Network
\item Target Network
\end{enumerate}

\par The target network receives the subsequent time state, while the main network receives the information about the present state. These networks’ outputs are input in the loss function, determining the error value. The target network’s parameters are synced with the main network’s parameters via the error gradient. Appropriate parameters are kept in a memory module named D for use when choosing the final state. The architecture can be summarized as follows:
\begin{itemize}
\item The input data is derived from the memory module D and includes the relevant state information represented by S, S’, and the quaternary state (S, a, r, S’).
\item  The input is transmitted simultaneously to the primary and reference networks. Real-time synchronization between these two networks ensures all their network settings are identical.
\item The LSTM unit receives the input status data and applies operations. These operations include processing the data using LSTM layers with an attention mechanism.
\item Subsequently, the data is flattened and passed to a fully connected layer. This layer comprises three individual layers, each containing 256 neurons.
\item The rectified linear unit (ReLU) activation function is employed throughout the network. 
\end{itemize}

\subsubsection{The Loss Function}
\par This study’s loss function utilizes a concept from DDQN. It involves two steps: 

\begin{enumerate}
    \item firstly, identifying the action that maximizes the Q-value, denoted as $max_{a} \mathcal{Q}(S, a’, \phi_{i})$, using the primary or main network parameters ($\phi_{i}$).
    \item Then, the corresponding action is determined in the target network. This Q-value is utilized to calculate the $\mathcal{Q}_{target}$ value.
\end{enumerate}

\par It’s essential to remember that the Q-value obtained from the target network may not necessarily be the highest, which helps prevent the selection of overestimated sub-optimal actions. We provide a mathematical representation of the loss function used in ARDDQN, which is expressed as:

\begin{equation}
Loss =\mathcal{E}\{(\mathcal{R}+\gamma \cdot \mathcal{Q}_{\text {target }}(s^{\prime},\arg \max \mathcal{Q}_{\text {main }}\left(s^{\prime}, a\right)-\mathcal{Q}_{\text {main }}(s, a))^2\}    
\end{equation}

\begin{algorithm}
\caption{DDQN-LSTM-Attention for Sequential Decision Making}
\begin{algorithmic}[1]
    \State Initialize DDQN neural network parameters: $\theta_{\text{DDQN}}$
    \State Initialize LSTM neural network parameters: $\theta_{\text{LSTM}}$
    \State Initialize Attention layer parameters: $\theta_{\text{Attention}}$
    \State Initialize DDQN target network parameters: $\theta_{\text{DDQN\_target}}$
    \State Initialize LSTM target network parameters: $\theta_{\text{LSTM\_target}}$
    \State Initialize experience replay buffer: $D$

    \For{episode = 1 to num\_episodes}
        \State Reset environment and initialize state: $s$
        \State Reset LSTM hidden states: $h, c$

        \For{t = 1 to max\_timesteps}
            \State Select action using epsilon-greedy policy based on DDQN and \indent\indent LSTM  predictions.
            \State Execute action, observe next state ($s'$), reward ($r$), and done flag \indent\indent($\text{done}$).

            \State Store ($s, a, r, s', \text{done}$) in experience replay buffer $D$.

            \State Sample random minibatch from $D$: minibatch
            \State Calculate target Q-values using DDQN and LSTM target networks:
                \State \quad $Q_{\text{DDQN\_target}} = \text{DDQN\_target}(s', a; \theta_{\text{DDQN\_target}})$
                \State \quad $Q_{\text{LSTM\_target}} = \text{LSTM\_target}(s', h, c, a; \theta_{\text{LSTM\_target}})$

            \State Apply attention mechanism to combine DDQN and LSTM outputs:
                \State \quad $Q_{\text{combined}} = \text{Attention}(Q_{\text{DDQN\_target}}, Q_{\text{LSTM\_target}}; \theta_{\text{Attention}})$

            \State Compute DDQN loss: $L_{\text{DDQN}} = \text{MSE}(Q_{\text{DDQN}}, Q_{\text{combined}})$
            \State Compute LSTM loss: $L_{\text{LSTM}} = \text{CustomLoss}(Q_{\text{LSTM}}, Q_{\text{combined}})$

            \State Update DDQN and LSTM networks using backpropagation:
                \State \quad $\theta_{\text{DDQN}} = \theta_{\text{DDQN}} - \alpha \nabla_{\theta_{\text{DDQN}}} L_{\text{DDQN}}$
                \State \quad $\theta_{\text{LSTM}} = \theta_{\text{LSTM}} - \beta \nabla_{\theta_{\text{LSTM}}} L_{\text{LSTM}}$

            \State Update target networks periodically:
                \State \quad if $t \mod \text{update\_target\_interval} == 0$:
                \State \quad \quad $\theta_{\text{DDQN\_target}} = \tau \theta_{\text{DDQN}} + (1 - \tau) \theta_{\text{DDQN\_target}}$
                \State \quad \quad $\theta_{\text{LSTM\_target}} = \tau \theta_{\text{LSTM}} + (1 - \tau) \theta_{\text{LSTM\_target}}$

            \State Update current state: $s = s'$

            \State Continue to the next timestep if the episode is not done.
        \EndFor
    \EndFor
\end{algorithmic}
\label{Algo:Alg1}
\end{algorithm}
\par In algorithm~\ref{Algo:Alg1}, initialize the parameters for the DDQN neural network, LSTM neural network, attention layer, and their corresponding target networks, and Set up an experience replay buffer (D) to store experiences for training. (lines 1-6). Reset the environment, initialize the state (s), and Reset the hidden states of the LSTM network (h and c) (lines 8-9). Select an action using an epsilon-greedy policy based on predictions from both the DDQN and LSTM networks (line 11). Execute the selected action and observe the next state (s'), the immediate reward (r), and a flag indicating if the episode is done (line 12). The tuple (s, a, r, s', done) in the experience replay buffer D is stored. The DDQN and LSTM target networks are used to calculate target Q-values for the next state (s') and combine the DDQN and LSTM outputs using an attention mechanism (lines 15-19). The loss for both DDQN and LSTM networks based on the combined Q-values is computed (lines 20-21). The parameters of the DDQN and LSTM networks using backpropagation and gradient descent are updated (lines 22-24). Periodically, the target networks are updated to make them slowly track the main networks (lines 25-28). The current state (s) to the next state (s') is updated.
\section{Experiments and Results}

\subsection{Experiment Setup}
\par The experiments were conducted using the following system specifications: an Intel(R) Xeon(R) Silver 4210 CPU operating at a frequency of 2.20GHz and a GeForce RTX 2080ti GPU running at a clock speed of 1.35GHz. The UAV operates in two distinct grid configurations. The first scenario, referred to as ``Manhattan32,” consists of a grid of cell size $32\times 32$, and in the upper left and bottom right corners, there are two takeoff and landing zones. In addition to the usual building patterns, this situation includes additional no-fly zones (NFZs) and buildings of odd shapes. The second scenario, ``Urban50”, encompasses a grid with dimensions of 50 × 50 cells. It features a single starting and landing area surrounding the central building. In this scenario, the buildings are often bigger and more dispersed. Moreover, the Map’s bottom portion has a second sizable NFZ. We assume each cell is $10m \times 10m $ in size in all scenarios.

\subsubsection{Coverage Path Planning}
\par A 90-degree field-of-view camera is mounted to the UAV, flying at a fixed height of 25 meters. The UAV may thus simultaneously cover a $5\times5$ cell area as long as nothing blocks its line of sight. The goal areas are developed by arbitrarily layering geometric objects of various sizes and kinds to produce target zones that are only loosely linked. The path length is a standard assessment statistic for the CPP problem. This statistic, however, only allows for a helpful comparison when complete coverage is feasible. In this study, our main objective is to examine flying time-limited CPP, where total coverage is often unachievable. The coverage ratio and collection ratio are, hence, the evaluation conditions used.

\subsubsection{Data Harvesting}
\par A UAV that interacts with devices on the ground while flying at a fixed height of 25 meters is the subject of the DH problem. The line-of-sight condition, distance, random shadow fading, and other variables affect the data rate that may be attained while utilizing the same communication channel settings that are explained~\cite{7}. The route length is not an important metric in this scenario, which is similar to CPP. According to the volume of data, the number of IoT devices, and the maximum flight duration, it is difficult to capture all the data in all cases. As a result, the collection ratio of data collected from all appliances to data originally accessible when each device is combined is used as an evaluation metric.

\subsection{Evaluation of local and global parameter}
\par To assess the performance of responsiveness to the new hyper-parameters,$l_s$ is used as local map scaling, and $g_s$ is used as global map scaling. We trained several agents under various conditions on both CPP and DH tasks. The four values we came up with for $l_s$ and $g_s$ for each conceivable combination were used to train three agents. Additionally, we trained agents only utilizing local map processing, corresponding to setting $g_s$= 1 and $l_s$= 0. This is also known as global map processing. The resultant agents completed 200k training steps before being put to the test 1000 Monte Carlo-generated scenarios. In this work, we consider the movement budget of 150-300. An agent with a global scaling parameter ($g_s$=3 or 5 ) and local scaling parameter ($l_s$=17) perform best regarding their flattened layer size. Hyperparameters used in this work are listed in Table~\ref{tab:table1}.

\begin{table}[ht]
\centering
\caption{Hyperparameters for Manhattan and urban scenarios}
\label{tab:table1}
\begin{tabular}{|c|l|c|c|}
\hline
\textbf{Parameters} & \textbf{Description}  & \textbf{Manhattan32} & \textbf{Urban50} \\ \hline

$\theta$             & Trainable Parameters         & 1,175,302            & 980,630    \\ \hline
$l_s$                & {local map scaling}          & 17                   & 17         \\ \hline
$g_s$                & {global map scaling}         & 5                    & 5          \\ \hline
n\textsubscript{c}   & No. of conv layers           & 2                    & 2          \\ \hline
n\textsubscript{k}   &  Kernel Size                 & 5                    & 5          \\ \hline
n\textsubscript{u}   &  RNN units                   & 16                   & 16         \\ \hline
bs                   & Batch size                   & 128                  & 128        \\ \hline
\end{tabular}
\end{table}

\subsection{Simulations}

\par
\begin{figure*}[ht!]
  \centering
  \includegraphics[width=\linewidth]{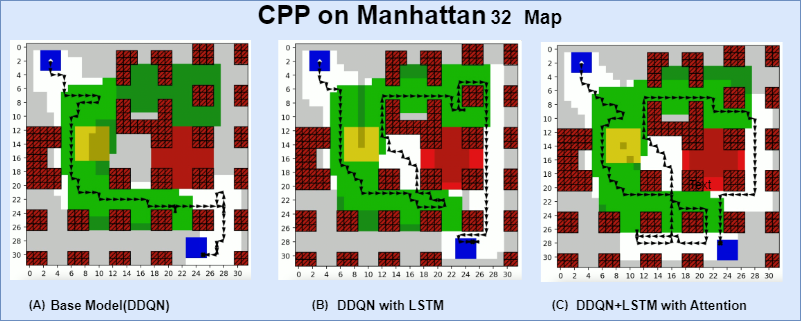}\hfill
  \caption{Coverage path planning on ``Manhattan32,"  Map}
  \label{fig:figure5}
\end{figure*}

 We considered two maps for simulation purposes in the coverage path planning scenario. The ``Manhattan32” map, which contains two takeoff and landing spots marked by blue and red, denotes the no-fly zone and is shown in Figure~\ref{fig:figure5}, the simulation of coverage route design. Here, we can see that a black line indicates the area that the UAV has covered while using the energy at its disposal. This shows that the base model (DDQN) covers a smaller region than the suggested model. The successful landing ratio is around 90\%, while the coverage ratio is about 55\%. DDQN produces superior landing and coverage ratios with LSTM while designing its coverage path. Our attention-based LSTM with the DDQN model delivers superior coverage compared to previous models.

\begin{figure*}[ht!]
  \centering
  \includegraphics[width=\linewidth]{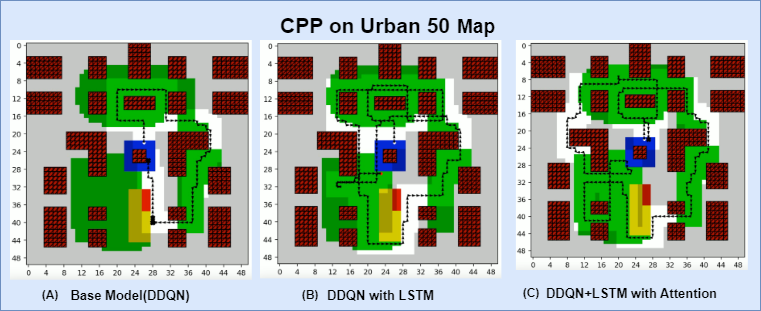}\hfill
  \caption{Coverage path planning on ``Urban50" Map}
  \label{fig:figure6}

\end{figure*}

For the sake of the experiment, we have taken into account the ``Urban50”, which has only one takeoff and landing point but a larger area as compared to ``Manhattan 32” in Figure~\ref{fig:figure6}, and we can see that our suggested model, an attention-based LSTM with DDQN, outperforms others.

\par
\begin{figure}
  \centering
 \includegraphics[width=7cm,height=8cm]{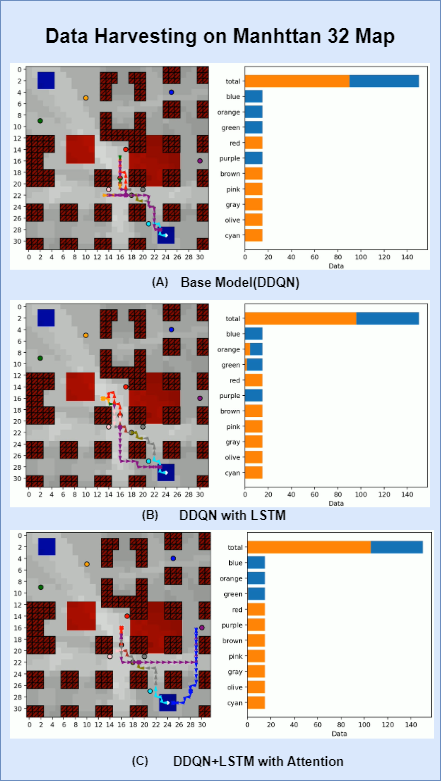}
  \caption{Data Harvesting on ``Manhattan32" Map}
  \label{fig:figure3}
\end{figure}

 We have considered the two maps for the data harvesting issues, similar to the coverage path problem. Here, the fundamental DDQN model completes its data collection in the ``Manhattan32” scenario at a rate of around 60\%. However, the UAV can access more IOT devices on the Map when the LSTM and attention mechanism are included in the network model. This considerably improves the collection and landing ratio, as shown in Figure~\ref{fig:figure3}. On the ``Manhattan32” map, in the data harvesting scenario, the UAV successfully landed with some data after collecting data from various IoT devices in the target area. These devices are indicated in different colors, such as green, red, and brown. Here, we can see that attention-based LSTM with DDQN can have more data than others.

\par
\begin{figure}
  \centering
  \includegraphics[width=7cm,height=8cm]{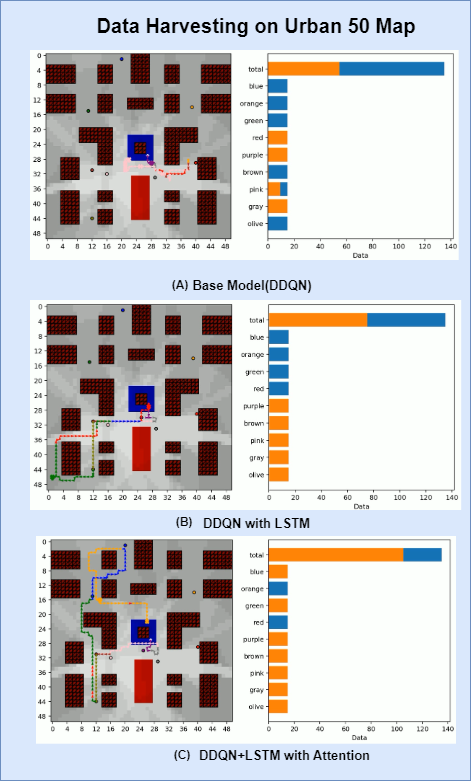}
  \caption{Data Harvesting on ``Urban50" Map}
  \label{fig:figure4}
\end{figure}

 Similarly, we tested the whole data-harvesting scenario using the large-scale map “Urban50” and obtained a data collection rate using a DDQN-based model. However, when we tried with a different model, such as LSTM, we got a greater collection ratio, as shown in Figure~\ref{fig:figure4}. We can see from this that the DDQN model and our suggested attention-based LSTM had an excellent collection ratio.

\subsection{Performance Evaluation}
\par The CPP agents received training in zones with 3 to 8 distinct forms; these shapes are denoted by various colours, such as blue, red, green, and others. The UAV’s starting and stopping points are shown by the colour blue, the NFZ is indicated by the colour red and the zone is indicated by the colour green by the absence of any obstacles. ``Manhattan32” and “Urban50” are within the purview of this zone identifier. When we consider the case of ``Manhattan32”, the agent may go up to 50-150 steps, while in the case of ``Urban50”, it may go up to 150-250 steps depending on the requirement, and UAV can occupy 20-50\% area of the entire Map. In the DH scenarios, ten devices with capacities ranging from 5.0 to 20.0 data units were dispersed randomly across open cells. The movement ranges were chosen at 50–150 steps and 100–200 steps for the ``Manhattan32” and ``Urban50”, respectively.

\par The agents showed they could find trajectories encompassing a significant amount of the target area in the CPP situations. The agents effectively covered most of the target region, including non-flight zones (NFZs), despite certain places that needed diversions being disregarded, leading to insufficient coverage. Even while other RNN models have improved the collection ratio and landing ratio, the attention-based LSTM mechanism-trained agents specifically showed a considerable increase in the collection ratio, coverage ratio, and landing ratio. There were 200k steps in the training for all agents. To test the agent performance for successful landing throughout the mission either in CPP or DH, 1000 Monte Carlo instances have been used. 

\begin{figure}
  \centering
  \includegraphics[width=0.7\linewidth]{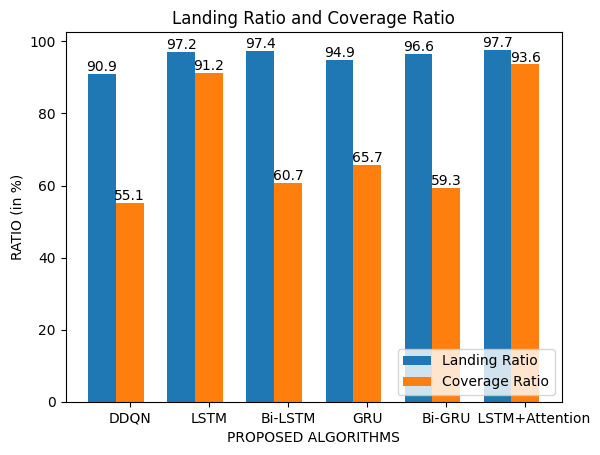}
  \caption{Coverage path planning on ``Manhattan32,"  Map}
  \label{fig:figure7}
\end{figure}

\begin{figure}
  \centering
  \includegraphics[width=0.7\linewidth]{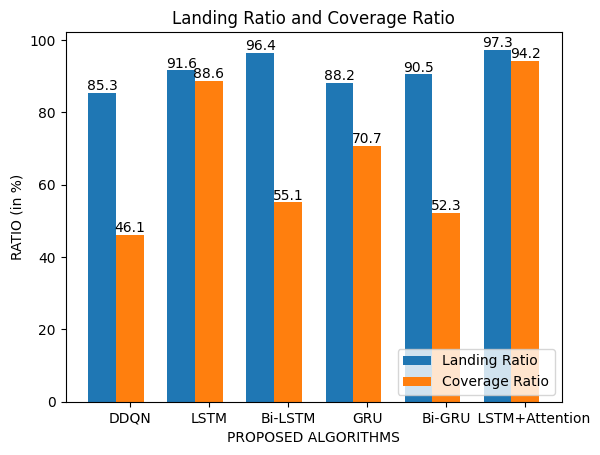}
  \caption{Coverage Path planning on ``Urban50" Map}
  \label{fig:figure8}
\end{figure}

\par In Figure~\ref{fig:figure7} and~\ref{fig:figure8}, the performance of various mechanisms is presented based on two maps, ``Manhattan32” and ``Urban50”, which assess coverage path planning based on the two-parameter landing ratio and coverage ratio. LSTM, Bi-LSTM, GRU, Bi-GRU, and LSTM with attention were some of the recurrent network algorithms we tested, and we compared them to the Base DDQN model. Compared with the model implemented without integrating recurrent neural networks into the double deep Q-Network, the model performance parameter collection ratio, coverage ratio, and landing ratio are evaluated based on the DDQN Network. Based on ``Manhattan32" map we got coverage ratio (55.1\%), landing ratio (90.9\%), and based on ``Urban50" we got coverage ratio (46.1\%), landing ratio (85.3\%). We proposed the model with R.N.N. integration that exceeds the landing ratio by 6.3\% (using LSTM), 6.5\% (using Bi-LSTM), 4.0\% (GRU), 5.7\% (using Bi-GRU), 6.8\% (using LSTM with attention) for the ``Manhattan32” Map scenario, and the landing ratio exceeds by 6.3\% (using LSTM), 11.1\% (using Bi-LSTM), 2.9\% (using GRU) 5.2\% (using Bi-GRU) and 12\% (LSTM with attention)for ``Urban50" Map scenario. Similarly, the coverage ratio also exceeds by 36.1\% (using LSTM), 5.6\% (using Bi-LSTM), 10.6\% (GRU), 4.2\% (using Bi-GRU), 38.5\% (LSTM with attention) for the ``Manhattan32,” Map. It exceeds the coverage ratio by 42.5\% (using LSTM), 9\% (using Bi-LSTM), 24.6\% (using GRU), 6.2\% (using Bi-GRU), and 48.1\%(using LSTM with attention)for the “Urban50” map scenario, so in the case of CPP our proposed model considerably outperforms better than the model which was implemented with the help of DDAN. The graph shows that LSTM with attention functions better than all others regarding landing and coverage ratios.

\begin{figure}
  \centering
  \includegraphics[width=0.7\linewidth]{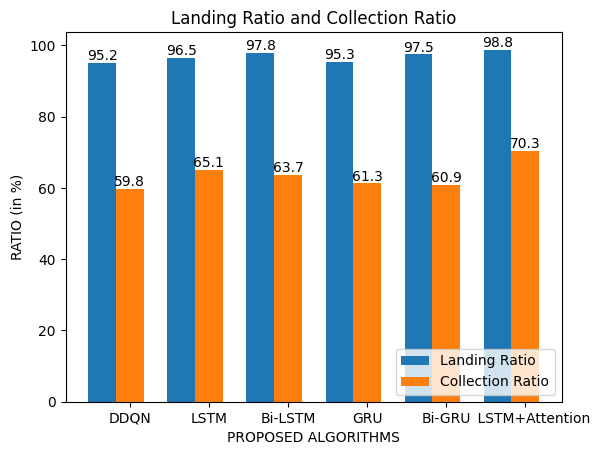}
  \caption{Data harvesting ``Manhattan32,"  Map}
  \label{fig:figure9}
\end{figure}

\begin{figure}
  \centering
  \includegraphics[width=0.7\linewidth]{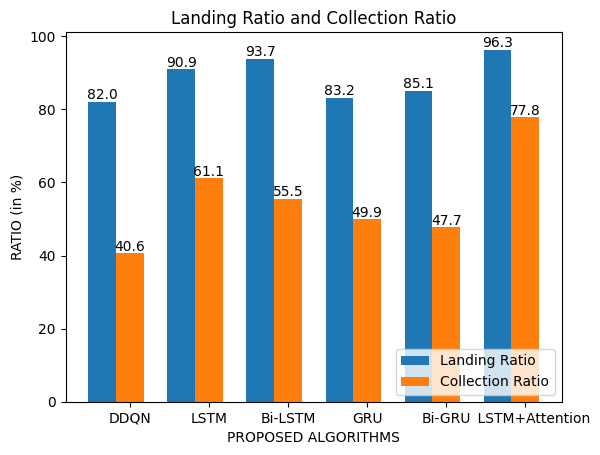}
  \caption{Data harvesting on ``Urban50" Map}
  \label{fig:figure10}
\end{figure}

\par In Figure~\ref{fig:figure9} and~\ref{fig:figure10}, the performance of various mechanisms is presented based on two maps, “Manhattan32” and “Urban50”, which assess data harvesting based on the two-parameter landing ratio and collecting ratio. We tested several recurrent network algorithms, such as LSTM, Bi-LSTM, GRU, Bi-GRU, and LSTM, with attention and compared the results to the Base (DDQN) network model. The model was implemented without integration of RNN with DDQN on two maps, the collection ratio (59.8\%) and landing ratio (95.2\%) for the “Manhattan32” Map, the collection ratio (40.6\%) and landing ratio (82.0\%) for the “Urban50” map respectively. Our proposed model exceeds the landing ratio by 1.3\% (using LSTM), 2.6\%(using Bi-LSTM), 0.1\% (using GRU), 2.3\% (using Bi-GRU), and 3.6\% (using LSTM with attention)for the “Manhattan32” map. It exceeds the landing ratio by 8.9\% (using LSTM), 11.7\% (using Bi-LSTM), 1.2\%(using GRU), 3.1\% (using Bi-GRU) and 14.3\% (using LSTM with attention) for the “Urban50” map. Similarly, Our proposed model also gives better performance in terms of collection ratio on “Manhattan32,” and exceeds by 5.3\% (using LSTM), 3.9\% (using Bi-LSTM), 2.5\% (using GRU), 1.1\% (using Bi-GRU), and 10.5\%(using LSTM with attention). The collection ratio on "Urban50" exceeds by 20.5 (using LSTM), 14.9\% (using Bi-LSTM), 9.3\% (using GRU), 7.1\% (using Bi-GRU), and 37.2\% (using LSTM with attention). So after observing this, we can say that our proposed model performs better in any of the cases as compared with existing DDQN without RNN integration, especially in the case of LSTM with the attention it is showing better performance in terms of landing ratio, collection ratio and coverage ratio in consideration with “Manhattan32” and “Urban50” map.

\section{CONCLUSION}
\par We have proposed the ARDDQN, a new approach for addressing critical challenges in Unmanned Aerial Vehicle (UAV) coverage path planning (CPP) and data harvesting (DH). We accomplish optimal path coverage decisions for effective data collecting from IoT devices by combining double deep Q-networks (DDQN) with recurrent neural networks (RNNs) and an attention mechanism. Long Short-Term Memory (LSTM) integration with the attention mechanism and a sophisticated global-local map processing technique considerably improves performance in various scenarios. Our emphasis on enhancing both landing and coverage rates in CPP and data collecting in DH demonstrates the efficacy of our suggested methodology. Consideration of multi-UAV systems as a future development promises even more robust data collection possibilities. 


\section*{Statements and Declarations}
\subsection*{Ethical Approval}
\par Not Applicable 

\subsection*{Availability of supporting data}
\par We will be available data from the corresponding author on reasonable request.

\subsection*{Competing interests}
\par Not Applicable

\subsection*{Funding}
\par Not Applicable

\subsection*{Authors' contributions}
\par Praveen Kumar: Conceptualization, Methodology, Experimentation, Writing: original draft; Priyadarshni: Writing: review and editing; Rajiv Misra: Conceptualization, Writing: review, Supervision.



\bibliography{sn-bibliography}%


\end{document}